\documentclass[sigconf, nonacm]{aamas} 
\usepackage{balance}

\newcommand\blfootnote[1]{%
  \begingroup
  \renewcommand\thefootnote{}\footnote{#1}%
  \addtocounter{footnote}{-1}%
  \endgroup
}

\title[Deep Interactive Bayesian Reinforcement Learning]{Deep Interactive Bayesian Reinforcement Learning \\ via Meta-Learning}

\author{Luisa Zintgraf$^*$}
\affiliation{\institution{University of Oxford}}

\author{Sam Devlin}
\affiliation{\institution{Microsoft Research}}

\author{Kamil Ciosek}
\affiliation{\institution{Microsoft Research}}

\author{Shimon Whiteson}
\affiliation{\institution{University of Oxford}}

\author{Katja Hofmann}
\affiliation{\institution{Microsoft Research}}

\begin{abstract}
Agents that interact with other agents often do not know a priori what the other agents' strategies are, but have to maximise their own online return \emph{while} interacting with and learning about others. 
The optimal adaptive behaviour under uncertainty over the other agents' strategies w.r.t.\ some prior can in principle be computed using the Interactive Bayesian Reinforcement Learning framework.
Unfortunately, doing so is intractable in most settings, and existing approximation methods are restricted to small tasks.
To overcome this, we propose to \emph{meta-learn} approximate belief inference and Bayes-optimal behaviour for a given prior.
To model beliefs over other agents, we combine sequential and hierarchical Variational Auto-Encoders, and meta-train this inference model alongside the policy.
We show empirically that our approach outperforms existing methods that use a model-free approach, sample from the approximate posterior, maintain memory-free models of others, or do not fully utilise the known structure of the environment.
\end{abstract}

\newcommand{\BibTeX}{\rm B\kern-.05em{\sc i\kern-.025em b}\kern-.08em\TeX}

\usepackage[utf8]{inputenc} 
\usepackage[T1]{fontenc}    
\usepackage{hyperref}       
\usepackage{url}            
\usepackage{booktabs}       
\usepackage{amsfonts}       
\usepackage{nicefrac}       
\usepackage{microtype}      
\usepackage{subcaption}
\usepackage{wrapfig,lipsum,booktabs}
\usepackage{multirow}
\usepackage{amsmath}
\usepackage{graphicx}
\usepackage{bm}
\newcommand\mydots{\hbox to 1em{.\hss.\hss.}}
\usepackage{color}
\definecolor{darkgreen}{rgb}{0.0, 0.4, 0.13}
\definecolor{orange}{rgb}{1,0.5,0}
\definecolor{greenish}{rgb}{0,0.9,0.5}

\newcommand\acro{MeLIBA}

\begin{document}

\pagestyle{fancy}
\fancyhead{}

\maketitle 

\blfootnote{
\small
$^*$ Work done during an internship at Microsoft Research Cambridge. \\
Correspondence: \url{luisa.zintgraf@cs.ox.ac.uk}, \url{katja.hofmann@microsoft.com}.
}


\section{Introduction}

Multiple agents that interact in the same environment influence each other through their actions: either directly by cooperating, communicating, or competing; or indirectly by affecting the state of the world. 
To successfully interact, such agents must learn about the other agents' strategies and reason about their behaviour.
For example, playing a game of soccer with a new team requires learning about each player's role and coordinating actions;
driving a car through busy traffic requires anticipating other drivers' moves and reacting to these appropriately; 
and 
successfully teaching a complex subject to a student requires adjusting the teaching method to their learning style.

A desirable capability of artificial agents that interact with other (human or artificial) agents is therefore the ability to adapt to others in an ad-hoc way, i.e., learn about the behaviour of other agents and adapt while doing so.
This implies that such agents must learn policies that maximise \textit{online return} incurred \textit{while} learning about initially unknown other agents.
When resources (data, time, and computation) for learning about others are limited, then key to adapting quickly is making best use of an agent's prior knowledge about the other agents' strategies. 
The better it can do this, the better it can learn about and adapt to others. 

A principled way to study the problem of adapting to unknown situations is from the perspective of decision-making under uncertainty.
An agent that acts optimally under uncertainty given some prior belief is called Bayes-optimal \citep{mitchell1997machine, duff2002optimal}, and the framework for computing such agents in a multi-agent setting is called Interactive Bayesian Reinforcement Learning \citep[IBRL; ][]{chalkiadakis2003coordination, hoang2013general}. 
In principle, this framework supplies the tools to compute an agent that acts optimally under uncertainty about other agents' strategies, and optimally trades off exploration (learning about others, possibly forfeiting large immediate expected rewards for higher long-term expected returns) and exploitation (taking promising actions in terms of expected return given the current model).
This approach requires maintaining a belief over the other agents' strategies, and computing the optimal action given that belief.
The IBRL framework is incredibly powerful, since its solution -- a Bayes-optimal policy -- yields agents that adapt in the best possible way to other agents.
Unfortunately, computing this solution is intractable for most interesting problems, and existing approximation methods are restricted to small environments and simple agent models \citep{chalkiadakis2007bayesian, nghia2014new}.

In this paper, we argue that the IBRL framework is a useful proxy for learning adaptive policies, and propose \textbf{Me}ta \textbf{L}earning \textbf{I}nteractive \textbf{B}ayesian \textbf{A}gents (\acro). 
We leverage recent advances in agent modelling, approximate variational inference, and meta-learning, to compute approximately Bayes-optimal agents for complex multi-agent settings in a general and tractable way.

Specifically, our model consists of two parts that are jointly meta-trained on a given prior distribution over other agents: 
(1) a belief inference network,
and 
(2) a policy that conditions on the approximate belief.
There is a rich history of modelling agents in the literature \citep[a.o.]{hula2015monte, oliehoek2016concise, albrecht2018autonomous}, and we build on ideas from \citet{rabinowitz2018machine} who use a hierarchical latent structure that separates permanent and temporal components of agent models.
To maintain \emph{beliefs} over these, we use a variational auto-encoder \citep[VAE,][]{kingma2013auto} for sequential data \citep{chung2015recurrent} combined with a hierarchical latent structure \citep{zhao2017learning}.
The policy conditions on the latent variables of this VAE, allowing it to adapt to others online.

Empirically, we demonstrate that 
explicitly learning and conditioning on an approximate belief over other agents' strategies can substantially improve performance in multi-agent settings, compared to relying on samples, or using a model-free policy with access to memory.
Our results indicate that factoring our model according to environment and the other agents' structure is beneficial when learning to interact with different agents, and that we can successfully learn adaptive policies.

\section{Background} \label{sec:background}

This section presents our problem setting, main assumptions,
and the IBRL solution framework.

\subsection{Environment}

We consider a Markov game \citep{shapley1953stochastic, van1981stochastic} with $N$ agents, defined as a tuple $M = (N, \mathcal{S}, \mathcal{U}, P_R, P_T, \gamma, H)$, 
where $\mathcal{U}=\mathcal{U}_1\times\dotsc,\times~\mathcal{U}_{N}$ is the joint action space, 
$\mathcal{S}$ the shared state space, 
$P_R(\textbf{r}_{t+1}|s_t, \mathbf{u}_t, s_{t+1})$ a reward function where $\textbf{r}\in\mathbb{R}^N$ and $\mathbf{u}_t=(u_1,\dotsc,u_{N})$ are the rewards and joint actions, 
and
$P_T(s_{t+1}|s_t, \mathbf{u}_t)$ are the state transition probabilities (for brevity this includes the initial state distribution $P_T(s_0)$).
At each timestep $t$ in state $s_t\in\mathcal{S}$, each agent $i$ takes an action $u^{(i)}_t\in\mathcal{U}$.
The actions of all agents are executed simultaneously, and each agent receives its own reward.

Each agent $i$ chooses actions according to its policy $a^{(i)}:\mathcal{S}\times\mathcal{U}\times\mathbb{R}\times\dotsc\times\mathcal{S}\rightarrow\mathcal{U}_i$ which maps interaction histories to action probabilities, $a^{(i)}(u_t|\tau^{(i)}_{:t})$.
Here we allow an agent to condition its actions not only on the current state, but on the history $\tau^{(i)}_{:t}=(s_0, \mathbf{u}_0, r^{(i)}_1, \dotsc, s_t)$ with full observability of states and other agents' actions, and its private reward.

\subsection{Objective}

We consider the task of learning a policy $a^{(1)}$ for agent $1$, with no control over the other agents' policies $\mathbf{a}^{(-1)}=a^{(2)}, \dotsc, a^{(N)}$.
Instead, we have a prior distribution over those policies
$P^{N-1}_A(\mathbf{a}^{(-1)})=\prod_{i=2}^N P_A(a^{(i)})$.
The objective of our agent is to perform well in expectation when faced with a random sample of policies from this distribution, within a single game:
\begin{equation} 
    \max_{a^{(1)}} ~~ \mathbb{E}_{P^{N-1}_A} \left[ \mathbb{E}_{P^\mathbf{a}_R, P^\mathbf{a}_T, \mathbf{a}} \left[ \sum_{t=0}^{H-1} \gamma^t r^{(1)}_{t+1} \right] \right] .
    \label{eq:main_objective}
\end{equation}
Since the other agents' policies are unknown to our agent, it has to simultaneously learn about the other agents, while adapting its behaviour accordingly in order to maximise its own \emph{online} return within the given horizon $H$.\footnote{We do not make assumptions about the other agents' rewards, but assume access to a reward $r^{(1)}$ that leads to the desired behaviour if it acts self-interestedly. This applies to cooperative, competitive, and mixed settings.}

\subsection{Interactive Bayesian RL} \label{sec:problem_setting:ibrl}

An agent that maximises Equation \eqref{eq:main_objective} acts optimally under uncertainty and is called \emph{Bayes-optimal}, assuming we treat the distribution  $P^{N-1}_A$ over other agents as our epistemic belief about the world.   
This agent maximises its expected online return by optimally trading off exploration (taking possibly costly but information-gathering actions) and exploitation (taking promising actions according to its current knowledge).
It gathers information about the other agents \textit{if and only if} this helps accumulate more rewards in expectation in the future \emph{within} the given horizon $H$, and adapts its strategy conditioned on its belief about the other agents.

In principle, a Bayes-optimal policy can be computed using the framework of Interactive Bayesian Reinforcement Learning (IBRL, \citealp{chalkiadakis2003coordination, hoang2013general}).
Given a prior belief over the other agents' strategies $P_A^{N-1}$, the objective in IBRL is to maximise the expected return \textit{under uncertainty}, shown in Equation \eqref{eq:main_objective}.
To this end, our agent maintains a belief about the other agents' strategies, the posterior distribution $p(\mathbf{a}^{(-1)}|\tau_{:t})$, where $\tau_{:t}=(s_0, \mathbf{u}_0, r^{(1)}_1, \dotsc, s_t)$ is the agent's experience until the current time step $t$. 
This posterior is updated deterministically at every timestep following Bayes' rule.

The agent's actions are now conditioned on hyper-states \citep{duff2002optimal} $s^+=(s_t, b_t)$: the environment states, together with the current belief.
We slightly abuse notation and write 
$b_t=b(\tau_{:t})=p(\mathbf{a}^{(-1)}|\tau_{:t})$ 
for the current belief, and condition the policy on this belief, denoted
\begin{equation} \label{eq:policy_conditioned_on_hyperstates}
    a^{(1)}(u_t|s_t, b_t) ~~~~~~~\text{  or  }~~~~~~~ a^{(1)}(u_t|s^+). 
\end{equation}
We assume a parameterised posterior where the belief $b_t$ is fully characterised by the distribution's parameters.

The transition and reward function for hyper-states are
\begin{equation}
    P_T(s^+_{t+1}| s^+_t, \textbf{u}_t) = 
     \mathbb{E}_{b_t}\left[P_T(s_{t+1}|s_t, \mathbf{u}_t)\right]  \delta(b_{t+1} {=} p(\mathbf{a}^{(-1)}|\tau_{:t+1}))
\end{equation}
where $\delta$ is the Dirac-delta function, and
\begin{equation}
   P_R(\textbf{r}_{t+1}| s^+_t, \textbf{u}_t, s^+_{t+1}) = \mathbb{E}_{b_t}\left[P_R(\textbf{r}_{t+1}|s_t, \mathbf{u}_t, s_{t+1})\right] .
\end{equation}

The policy that maximises the expected return in the resulting belief MDP \citep{kaelbling1998planning} is the policy that optimally adapts to other agents given some prior belief about their strategies.
Unfortunately, computing this solution analytically is intractable for all but the smallest tasks, and even existing approximation methods are restricted to small environments.
In Section \ref{sec:meliba}, we present a method that aims to overcome this limitation.

\subsection{Prior Beliefs over Other Agents} \label{sec:background:prior}

We assume that the support of the prior distribution $P_A^{N-1}$ is over agents of the general form $a^{(i)}(u_t|\tau^{(i)}_{:t})$ described above.
These agents can be non-stationary in that they can adapt their behaviour depending on the interaction history within a \emph{single} game, up until the horizon $H$.
For example, an agent might decide whether to cooperate or not based on how often the other agents cooperated.
Any general solution therefore has to support these general types of other agents.

While this means that agents can learn within an episode, we do not model learning across episodes. 
At the beginning of each game, when sampling new agents from this prior, the other agents' memory is reset ($\tau_0=\emptyset$).
Modelling other agents that learn across episodes, not only within episodes, is in principle possible by adding a prior over the other agents' learning algorithm and changing Equation \eqref{eq:main_objective} to also incorporate this prior over the other agents' lifetime (not only until the episode horizon $H$).
This is however makes belief inference and planning \emph{even} harder and is beyond the scope of this paper.
We discuss this potential future direction in Section \ref{sec:discussion}.

In our approach, we use the prior distribution $P_A^{N-1}$ for sampling the other agents' policies i.i.d.\ during meta-training (see Section \ref{sec:meliba}).

\section{Related work} \label{sec:related_work}

Adapting to unknown other agents is crucial for many applications where agents can be deployed, and their limited ability to do so is a critical limiting factor for real-world applications.
For example, in ad-hoc teamwork \citep{stone2010ad}, agents trained to coordinate with each other fail to do so when paired with unseen partners \citep{carroll2019utility, canaan2020evaluating}.
We argue that a desirable objective for adaptive agents is the expected \emph{online} return, and therefore we should aim to learn approximately Bayes-optimal behaviour.

The notion of Bayes-optimality in RL was used by \citet{duff2002optimal} when formulating the Bayes-adaptive Markov Decision (BAMDP) process for single-agent MDPs with unknown reward and transition dynamics.
Like in the IBRL framework, belief inference and planning in belief space is generally intractable in BAMDPs, but recently meta-learning has been proposed as a more scalable way to compute approximately Bayes-optimal agents \citep{ortega2019meta, humplik2019meta, zintgraf2020varibad}.
To use the same ideas in a multi-agent setting, we could simply model other agents as part of the environment \citep[Multiple Individual Learners;][]{tan1997multi} and maintain a belief over the reward and transition function.
However, ignoring the known structure in the environment conflates several sources of stochasticity, making learning harder. 
By contrast, in the IBRL formulation (see Sec \ref{sec:problem_setting:ibrl}), we explicitly model beliefs over the other \emph{agents}.
In principle, this provides the right structure to solve the problem -- however, computing the solution is generally intractable and existing work provides approximate solutions restricted to small environments or restrictive assumptions \citep{chalkiadakis2007bayesian, hoang2013general, chalkiadakis2010cooperative, nghia2014new}.

Related to IBRL, a large body of literature on Bayesian RL for multi agent settings and game theory uses beliefs over other agents, e.g.,
to compute best responses, maximise value gain, or learn models that can be used for planning \citep[a.o.,][]{carmel1990exploration, nachbar2005beliefs, albrecht2016belief, sadigh2016information}.
For a general introduction to Bayesian reinforcement learning see \citet{ghavamzadeh2016bayesian}.
What separates \acro~from some of these approaches is that we optimise for \emph{Bayes-}optimal behaviour, i.e., we care about the online return the agent achieves \emph{while} learning about the other agents, as opposed to optimal behaviour that generally cannot be executed from the first timestep because the other agents' strategies are still unknown.

A requirement for IBRL is the need to maintain a distribution over other agents, and to do so explicitly we need to model them. 
We outline the most relevant agent modelling literature here, and refer to \citet{albrecht2018autonomous} and \citet{hernandez2017survey} for comprehensive surveys.
A popular way of modelling other agents is type-based modelling which assumes that the other agent has one of several types that can be pre-defined by an expert, or learned from data \citep{albrecht2019reasoning, barrett2013teamwork, stone2010ad}.
In this paper, we learn a latent variable model, in which the latent variables can be seen as a continuous representation of agent types learned in an unsupervised way via interaction.
Deep representation learning has been used in multi-agent settings by \cite{grover2018learning}, albeit not for online adaptation.
A central concept we use is separating behaviour shared by all other agents, and agent-specific permanent and temporal features. 
This is based on ideas from \citet{rabinowitz2018machine}, who refer to these as the agent's character (which is fixed throughout its life), and its mental state (which can change at any time and includes beliefs over other agents).
\citet{rabinowitz2018machine} consider a purely observational setting, whereas we are interested in learning Bayes-optimal behaviour in an \emph{interactive} setting.
This requires maintaining beliefs over the other agents' models, for which we combine hierarchical and sequential VAEs \citep{kingma2013auto, chung2015recurrent, zhao2017learning} to meta-train a belief inference model.
Further, it requires training a policy that interacts with (and influences) other agents, and uses these beliefs when making decisions to maximise online return.

Closely related to our approach is the work of \citet{papoudakis2020variational}, who consider a similar problem setting to the one outlined in Sec \ref{sec:background}.
They focus on partially observable settings, and approximating global from local (agent-specific) information.
To this end, they use a VAE to model other agents, and train an agent to use the latent variable to adapt.
In contrast to our setting, they assume the other agents are Markov and only condition their actions on the current state -- i.e., the other agents do not themselves adapt to the agents around them.
Instead, we allow full observability but focus on the question of how to model other agents that are non-stationary.
We show empirically that a different type of model is necessary if such non-stationary policies are allowed.
Most importantly, \citet{papoudakis2020variational} condition their policy on a \emph{sample} from the approximate belief.
However, this means they cannot learn a Bayes-optimal policy, since the agent cannot take into account its uncertainty about the other agents' policies.
Like \citet{papoudakis2020variational}, many existing methods consider the other agents to be Markov, i.e., their policy depends only on the current state \citep{he2016opponent, carroll2019utility, he2016opponent}.
Our model in contrast can model agents that condition their actions on a history, and adapt to non-Markov agents.

Recently, \citet{smith2020learning} propose Q-Mixing, which uses transfer learning to transfer Q-values to new mixture of opponents, and an opponent classifier to refine the mixing of Q-values to adapt to the new mix of opponents.
In comparison, our method does not rely on having a discrete (small) number of other agents, but can also represent continuous distributions over other agents' policies.

While we use an explicit approximate belief over other agents, an alternative from the meta-learning literature is a model-free version, where our agent is represented by a recurrent network and conditions its actions on the interaction history (including states, actions, and rewards) \citep{duan1611rl2, wang2016learning}.
Adapting to others is possible due to the recurrent structure of the model. 
Previous work shows that modelling other agents as auxiliary tasks in multi-agent settings can help learning \citep{hernandez2019agent}.
Similarly, we observe that the model-free approach performs worse than explicitly using the inductive bias of what to model (the other agent's actions).

Several methods have also been developed for adaptive agents on specific domains or making use of the structure of certain games, such as Avalon \citep{serrino2019finding}, Hanabi \citep{foerster2018bayesian, hu2019simplified, canaan2020generating}, or Poker \citep{bard2007particle, southey2012bayes}.
Our model on the other hand is general and can be applied to cooperative and competitive settings, and mixed and general sum games.

\section{Meta Learning Interactive Bayesian Agents} \label{sec:meliba}

\begin{figure*}[h]
    \centering
    \includegraphics[width=0.7\textwidth]{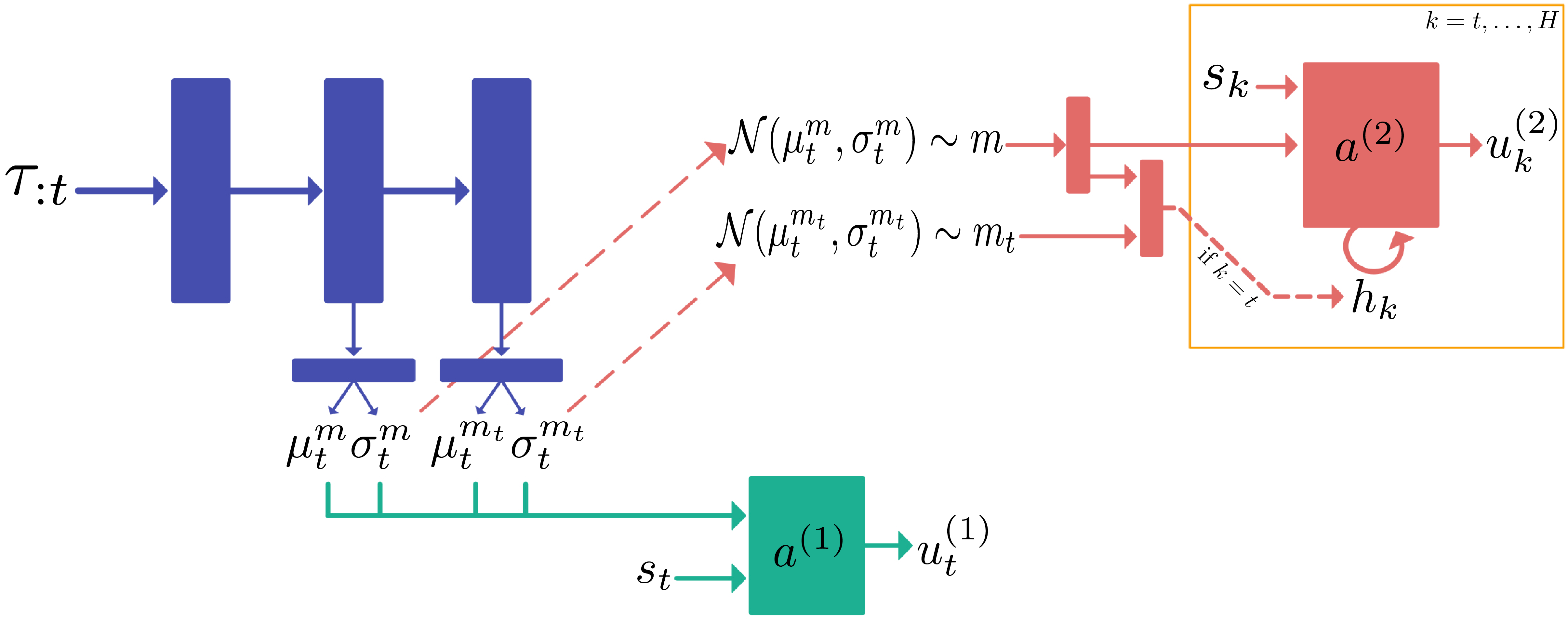}
    \caption{\acro~Network Architecture (see Sec \ref{sec:meliba:components}).}
    \label{fig:architecture}
\end{figure*}

In this section we introduce \textbf{Me}ta \textbf{L}earning \textbf{I}nteractive \textbf{B}ayesian \textbf{A}gents (\acro), a method for meta-learning approximately Bayes-optimal agents that adapt to other agents. 
To this end, we bring together the theoretical motivation from Interactive Bayesian RL with modern methods from meta learning, agent modelling, and belief inference.
In particular, we propose a hierarchical sequential VAE to model beliefs over other agents, and jointly train this inference network with a policy that conditions on environment states and approximate beliefs.

\subsection{Modelling Other Agents}

If we were in a tabular setting, with stationary other agents and a tractable prior and likelihood over their policies,
we could simply compute posterior distribution over state-action probabilities per other agent using Bayes' rule,
and use this to represent the belief over the other agents' policies.
We are however interested in more complex (often continuous) environments, other agents that are potentially non-stationary (in the sense that they condition their actions on the within-game interaction history), and we want to be able to deal with arbitrary priors.
Therefore, we learn a more complex belief representation using deep learning.

We model each other agent by its own permanent latent variable ($m$, also called \emph{agent character}) and a temporal latent variable ($m_t$, also called \emph{mental state}):
\begin{equation}
    a^{(i)}(u_t|\tau_{:t}) \equiv a(u_t|s_t, m^{(i)}, m_t^{(i)}).
\end{equation}
The character $m$ does not change throughout the agent's lifetime. 
The mental state $m_t$ can change in response to new observations at every timestep and allows us to model agents with non-stationary policies, i.e., policies conditioned on the interaction history.
For example, the mental state of an agent can represent counts of how often other agents have cooperated, or capture the other agents' belief over other agents (and their beliefs, and so on).

This can be viewed as a probabilistic extension of an idea by \citet{rabinowitz2018machine}, who propose this split to model other agents and coined the terms agent character and mental state. 
Compared to \citet{rabinowitz2018machine} who model agents in a purely observational setting, we want to learn to \textit{interact} with other agents, and therefore need to maintain beliefs over the other agents.
That is, we want to infer the posterior $p(\mathbf{m}, \mathbf{m}_t|\tau_{:t})$ given the agent's experience $\tau_{:t}$ up until the current timestep,
where $\mathbf{m}=(m^{(2)}, \dotsc,m^{(N)})$ and $\mathbf{m}_t=(m_t^{(2)}, \dotsc, m_t^{(N)})$.
We do so using a hierarchical sequential VAE, described next.

\subsection{Approximate Belief Inference}

Our agent's objective is to maximise expected future return (Eq \eqref{eq:main_objective}), the expected return given its current belief over the other agents' policies. 
Our agent therefore must predict other agents' future behaviour. 
We use this fact as an inductive bias to our model during meta-training, by learning to predict the future actions $p(u^{(i)}_{t:H}|s_t)$ of each of the other agents $i$ at every timestep $t$.

Optimising $p(\mathbf{u}_{t:H}|s_t)$ is intractable. Instead, we can optimise an evidence lower bound (ELBO):
\begin{align}
    \log ~ p(\mathbf{u}_{t:H}|s_t)  
    & \ge 
    ~ \mathbb{E}_{q(\mathbf{m}, \mathbf{m_t}|\tau_{:t})} \left[ \log g^\text{act} \right]
    \label{eq:elbo_pt1}
    \\
    & 
    \hspace{0.4cm} 
    - KL(q(\mathbf{m}, \mathbf{m_t}|\tau_{:t})||q(\mathbf{m}, \mathbf{m}_t|\tau_{:t-1}))
    \label{eq:elbo_pt2}
    \\
    & = ~ELBO_t ,
\end{align}
where
\begin{equation} \label{eq:ebo_pt1_extra}
    g^\text{act} = \mathbb{E}_{p(\mathbf{u}^{(-i)}_{t:H-1}, s_{t+1:H} | \mathbf{u}_t, s_t, \mathbf{m}, \mathbf{m_t})} 
    \Bigg[ 
    \prod_{k=t+1}^{H-1} p(\mathbf{u}_k | s_k, \mathbf{m}, \mathbf{m_k} )  
    \Bigg]
    .
\end{equation}
Our model is a special type of VAE \citep{kingma2013auto}, and combines a sequential VAE \citep{chung2015recurrent} with a hierarchical latent structure \citep{zhao2017learning}.
We provide an intuitive explanation of this objective here, and the full derivation is given in Appendix \ref{appendix:elbo_derivation}.

Line \eqref{eq:elbo_pt1} (RHS) / Eq \eqref{eq:ebo_pt1_extra}: 
At timestep $t$, given the current posterior $q(\mathbf{m}, \mathbf{m_t}|\tau_{:t})$, we predict the other agents' future actions $p(\mathbf{u_k} | s_k, \mathbf{m}, \mathbf{m_k})$ for all future timesteps $k=t+1,\dotsc,H$.
Since we assume independence between agents, this term factors across the other agents.
During meta-training, we have access to these future actions from the collected rollouts.
Future actions depend on the mental state $m^{(i)}_k$ for each other agent $i$, which changes at each timestep $k$ in the future.
Therefore, the model we maintain of the other agent includes a latent variable $m_k$ that evolves over time, for which the encoder outputs only the current mental state, $m_t$.
We model this using a recurrent architecture.

Line \eqref{eq:elbo_pt2}:
We use a hierarchical latent structure where the agent's temporal state $m_t$ can depend on its permanent type $m$.
We use a \emph{single} latent variable model for this, i.e., we do not factor the posterior distribution.
This follows insights by \citet{zhao2017learning} who show (for non-sequential VAEs) that learning hierarchical features can be obtained by having a single latent variable model, but a hierarchical structure in the network architecture.
Intuitively, this means that we generate $\mathbf{m}$ from early layers in the encoder and $\mathbf{m}_t$ from deeper layers. Similarly, the decoder has a reverse hierarchical structure (see Fig \ref{fig:architecture}).
Following \cite{zintgraf2020varibad}, we choose the prior (right-hand inside the KL) to be the previous posterior, akin to a filtering-type Bayesian update. 
This incentivises the posterior distribution to change slowly over time as the agent collects more data.

In practice, we represent the posterior using a Gaussian distribution $\mathcal{N}(\boldsymbol\mu_t, \boldsymbol\sigma_t)$, 
where $\boldsymbol\mu_t=(\mu_t^{m, (2)}, \mu_t^{m_t, (2)}, \dotsc, \mu_t^{m, (N)}, \mu_t^{m_t, (N)})$
and $\boldsymbol\sigma_t=(\sigma_t^{m, (2)}, \sigma_t^{m_t, (2)}, \dotsc, \sigma_t^{m, (N)}, \sigma_t^{m_t, (N)})$.
For the prior for the first timestep we set $q(\mathbf{m}, \mathbf{m}_t)=\mathcal{N}(\mathbf{0}, \mathbf{1})$.

A Gaussian distribution is a frequent choice in the VAE literature \citep{kingma2013auto} due to its favourable properties for training and sampling, and because it can represent complex distributions of observations due to the non-linearity in the encoder. 

\subsection{Meta-Learning Bayes-Adaptive Policies}

Given the approximate posterior $q(\mathbf{m}, \mathbf{m}_t|\tau_{:t})$, we want to learn an approximately Bayes-optimal policy. 
To this end, we condition our policy not only on the environment state, but also on this approximate belief over the other agents' policies (see Equation \eqref{eq:policy_conditioned_on_hyperstates}).
This enables approximately Bayes-optimal behaviour: the policy can take into account its uncertainty over the other agents' policies when choosing actions, and use it to trade off exploration and exploitation.
In practice, we approximate $q$ using a Gaussian distribution $\mathcal{N}(\bm\mu, \bm\sigma)$. 
The approximate posterior is fully characterised by the mean $\boldsymbol\mu_t$ and variance $\boldsymbol\sigma_t$.
The policy is then trained using standard RL methods by conditioning on environment states $s_t$ and beliefs $b_t=(\boldsymbol\mu_t, \boldsymbol\sigma_t)$, 
$a^{(i)}(u_t|s_t, \bm\mu_t, \bm\sigma_t)$.

\subsection{Putting Everything Together} \label{sec:meliba:components}

We use deep neural networks to represent 
\begin{itemize}
    \item \textcolor{blue}{\textbf{An encoder}} $b_\phi(\tau_{:t})=(\bm\mu, \bm\sigma)$ parameterised by $\phi$, where the outputs fully characterise the approximate posterior $q(\mathbf{m}, \mathbf{m}_t|\tau_:t)=\mathcal{N}(\mathbf{m}, \mathbf{m}_t|\bm\mu, \bm\sigma \mathbb{I})$. 
    The encoder has a hierarchical structure (see Fig \ref{fig:architecture}). 
    Each agent's latent variables $(m, m_t)$ are modelled using a diagonal Gaussian of size $M+M_t$, such that $\mu^{m, (i)}, \sigma^{m, (i)} \in \mathbb{R}^M$ and $\mu^{m_t, (i)}, \sigma^{m_t, (i)} \in \mathbb{R}^{M_t}$.
    \item \textcolor{red}{\textbf{A recurrent action decoder}} to predict the other agents' actions, $a^{(i)}_\theta(u_t|s_t, m^{(i)}, m_t^{(i)})$, parameterised by $\theta$.
    Since there are several other agents, the encoder outputs $N$ distribution parameters, 
    $\bm{\mu}=(\mu^{(2)},\dotsc,\mu^{(N)})$ and $\bm{\sigma}=(\sigma^{(2)},\dotsc,\sigma^{(N)})$, 
    which are independently fed into the same decoder to compute the reconstruction loss.
    \item \textcolor{darkgreen}{\textbf{A policy}} $a^{(1)}_\psi(u_t|s_t,b_t)$, parameterised by $\psi$ and dependent on $\phi$ (shorthand $a^{(1)}_{\psi,\phi}$).
\end{itemize}
Fig \ref{fig:architecture} shows the network architecture (assuming \emph{one} other agent), with the colours above corresponding to the colours in the figure.

The overall objective of our agent is to maximise
\begin{equation}
    \mathcal{L}(\phi, \theta, \psi) = \mathbb{E}_{P_A^{N-1}} \left[ \mathcal{J}(\phi, \psi) + \lambda \sum_{t=0}^{H-1} ELBO_t(\phi, \theta) \right], 
\end{equation}
where 
$
    \mathcal{J}(\phi, \psi) = \mathbb{E}_{P^\mathbf{a}_R, P^\mathbf{a}_T, a^{(1)}_{\psi,\phi}} \left[ \sum_{t=0}^T \gamma^t r_t \right] .
$
In our experiments we train the policy using PPO, and do \textit{not} backpropagate the RL-loss through the encoder. 
This saves computation, and allows us to investigate whether learning to predict the other agents' future actions from the latent structure of a VAE is sufficient to learn Bayes-optimal behaviour.
We therefore alternate between updating the VAE using the ELBO, and updating the agent using the RL loss.
For experiment details and hyperparameters see the appendix.

\section{Empirical Evaluation} \label{sec:experiments}

In this section we study \acro~and our modelling choices empirically.
We first look at a small matrix game that allows us to analyse the hierarchical structure that emerges in the latent space.
We compare \acro~to existing approaches for adaptive policies in complex multi-agent systems:
\begin{itemize}
    \item RL$^2$ \citep{duan1611rl2, wang2016learning}, which is a model-free meta-learning method. The architecture is therefore similar to Figure \ref{fig:architecture} but with no decoder and no hierarchy in the encoder (no $m_t$). Unlike in \acro, the RL loss is backpropagated through the encoder.
    \item LIOM \citep{papoudakis2020variational}. The main differences to \acro~are that the policy receives a \emph{sample} from the approximate posterior, the encoder is not hierarchical (no $m_t$), and the decoder is not recurrent.
    \item An \emph{average} policy that cannot adapt but learns a policy that is good on average across all other agents.
\end{itemize}
\acro~and all baselines are based on PPO (for further implementation details see the appendix).

We also study our architecture choices quantitatively and qualitatively by ablating the hierarchical and the sequential structure.

\subsection{Game of Chicken} \label{sec:experiments:chicken}
 
\setlength{\tabcolsep}{2pt}
\begin{wraptable}{r}{4cm}
\centering
    \begin{tabular}{ccccc}
    \cline{2-3}
    \multicolumn{1}{c|}{}   & \multicolumn{1}{c|}{Swerve}      & \multicolumn{1}{c|}{Straight}      &  &  \\ \cline{1-3}
    \multicolumn{1}{|c|}{Swerve} & \multicolumn{1}{c|}{(1, 1)} & \multicolumn{1}{c|}{(0, 8)} &  &  \\ \cline{1-3}
    \multicolumn{1}{|c|}{Straight} & \multicolumn{1}{c|}{(8, 0)} & \multicolumn{1}{c|}{(-1, -1)} &  &  \\ \cline{1-3}
                            &                             &                             &  & 
    \end{tabular}        
\caption*{Game of Chicken}      
\label{fig:chicken_results:matrix}
\end{wraptable} 

We consider a simple matrix game, Game of Chicken, because it allows us to verify that \acro~works as expected on a game where we know the solution and can easily analyse and illustrate the properties of \acro, such as the learned latent structure and some architecture choices. 

\begin{figure*}
\centering
\begin{subfigure}{\columnwidth}
    \centering
        \includegraphics[width=\columnwidth]{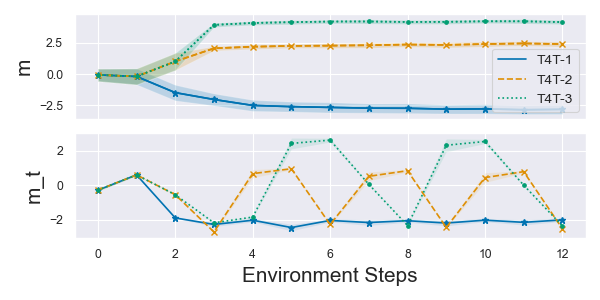}
        \caption{Latent variables during test rollout.}
        \label{fig:chicken_results:latent_visualisation}
\end{subfigure}
\hspace{2cm}
\begin{subfigure}{0.75\columnwidth}
    \centering
    \includegraphics[width=\columnwidth]{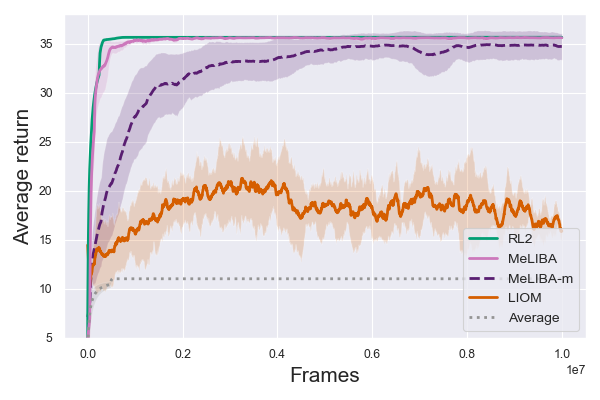}
    \caption{Learning Curves}
    \label{fig:chicken_results:learning_curve}
\end{subfigure}
\caption{Results for the Game of Chicken. (a) The latent variables of \acro~during three separate rollouts at test time, when playing against the different types of other agents T4T-1/2/3. Each colour represents a different game against one of these agents. The top shows the permanent latent variable $m$, which clearly separates the different agent types. The bottom the temporal latent variable $m_t$ which keeps track of how often the agents have cooperated. (b) The learning curves for \acro~and ablations/baselines. Both RL$2$ and \acro~learn the Bayes-optimal behaviour quickly. Using only a permanent latent state (\acro-m) leads to a performance drop because the other agent is not modelled appropriately. LIOM, which conditions the policy on \emph{samples} of the latent variable, underperforms in this environment, possibly because of the small latent dimension combined with the noise introduced via sampling.}
\label{fig:chicken}
\end{figure*}

\textbf{Environment Set-up.}
In the Game of Chicken \citep{bergstrom1998evolution} is a $2$-player competitive game.
Imagine two cars driving towards each other: if nobody swerves, they crash and get a penalty ($-1$); if they both swerve they get a medium reward ($1$); and if only one of them swerves they are "the chicken" and get a low reward ($0$) and the other player gets a high reward ($8$).
We hand-code three Tit-4-Tat agents, a strategy commonly used in game theory \citep{heap2004game} which swerves if the opponent swerved once / twice / three times in a row (T4T-1/2/3).
We randomly sample an agent to play with for $13$ repetitions, which is long enough that the optimal strategy requires inferring and remembering the opponent's strategy, and short enough to analyse game play.
As this is a stateless game, the only way to infer information about the other agent is to consider the actions it takes.
The Bayes-optimal strategy is to swerve until the other agent also swerves and thereby reveals which T4T strategy it is using, after which it can be exploited by swerving just until the other agent will swerve, and then going straight to get a payoff of $8$. 

\textbf{Latent Visualisation.}
For \acro~we use a latent dimensionality of $1$ each for the permanent and temporal aspects, $m\in\mathbb{R}$ and $m_t\in\mathbb{R}$ so we can analyse the latent space qualitatively.
Figure \ref{fig:chicken_results:latent_visualisation} shows how the latent variables behave when we roll out the learned (Bayes-optimal) policy.
The top is the mean of the latent variable $m$, plus its learned standard deviation (shaded areas). 
The separation between agent types happens after just $1$-$2$ timesteps, which is how long it takes before the other agent starts swerving (given that \acro~learned to always swerve at the beginning).
As expected, the standard deviation of the learned latent belief is high at the beginning of the episode and goes down as the agent gets more information about the other agent.
At the bottom of Figure \ref{fig:chicken_results:latent_visualisation} we show the same visualisation for the temporal latent variable, $m_t$.
This shows that the model learned to count the number of swerves, which is the behaviour we intended.
In Appendix \ref{appendix:results:chicken} Figure \ref{fig:appendix:chicken_rollouts}, we show the latent variables of other architectures (no hierarchy, or only permanent/temporal latents) 
which do not give the same desired latent structure. 

\textbf{Performance Comparison.}
Figure \ref{fig:chicken_results:learning_curve} shows the performance of \acro~, compared to several baselines. 
RL$^2$ (trained with a $128$-dimensional latent variable where \acro~has a $2$D bottleneck) learns to solve the task quickly.
This is unsurprising given the simplicity of the game and because the policy conditions directly on the entire interaction history via the recurrent encoder.
LIOM however performs poorly on this task (with similar architecture as \acro).
We believe that the reason is that the latent space is so small, and small perturbations (caused by the sampling) make PPO unstable.
We found that backpropagating the RL loss through the encoder, or increasing the latent dimension, helped to some extent (see Figure \ref{fig:appendix:chicken_learning_curve} in the Appendix).
But even then, LIOM does not learn to solve the problem. This can be explained by the fact that LIOM does not have the appropriate agent model (i.e., decoder) for opponents that condition on the interaction history. 
We test this hypothesis by testing \acro~with \emph{only} the fixed latent $m$ and a feed-forward decoder (denoted \acro-m).
As Figure \ref{fig:chicken_results:learning_curve} shows, \acro-m cannot solve the task given the wrong model for the other agent.
However it does outperform LIOM by a large margin, indicating that sampling the latent variable for the policy in LIOM, as opposed to conditioning on the entire posterior, causes LIOM's poor performance in this environment.

\subsection{Treasure Hunt} \label{sec:experiments:treasure}

In this section we focus on whether predictive structure improves over model-free approaches.
A model-free approach, where the policy is a recurrent network conditioned on its trajectory as in RL$^2$, can in principle learn the Bayes-optimal strategy \cite{ortega2019meta}.
However in practice we expect it to be difficult to learn to both infer the other agents' type, and use this information in the right way.
In most settings studied in the single-agent setting, there is a close mapping between tasks and reward, for example if the agent gets rewarded when it reaches a goal, or if the reward is a dense signal of which direction the agent has to walk, and that the right inductive biases can help.
In some multi-agent settings however, the mapping from task (i.e., what the other agent's strategies are) and the reward which the agent observes in the environment is more complex.
The reason is that the reward function depends on the other agent's actions, which in turn can depend on complex history-dependent strategies.
We therefore hypothesise that meta-learning to perform inference of the other agent's strategies is useful for maximising expected online return, 
and that adding the inductive bias of predicting the other agent's future actions like in \acro~can therefore help learning.

\begin{figure}
\begin{minipage}{\columnwidth}
    \begin{subfigure}{\columnwidth}
    \centering
    \begin{tabular}{ccccc}
    \hline
    \multicolumn{1}{|l|}{~MeLIBA~} & 
    \multicolumn{1}{l|}{~MeLIBA-m~} & 
    \multicolumn{1}{l|}{~LIOM~} &
    \multicolumn{1}{l|}{~RL$^2$~} & 
    \multicolumn{1}{l|}{~Average~} 
    \\ \hline
    \multicolumn{1}{|c|}{\textbf{10.1} (0.1)} & 
    \multicolumn{1}{c|}{\textbf{10.0} (0.2)} & 
    \multicolumn{1}{c|}{\textbf{10.2} (0.05)} & 
    \multicolumn{1}{c|}{7.7 (0.3)} & 
    \multicolumn{1}{c|}{1.3 (0.9)} 
    \\ \hline       
    \end{tabular}
    \caption{Meta-Test Performance}
    \end{subfigure}
    \begin{subfigure}[b]{0.4\columnwidth}
        \centering
        \includegraphics[width=\columnwidth]{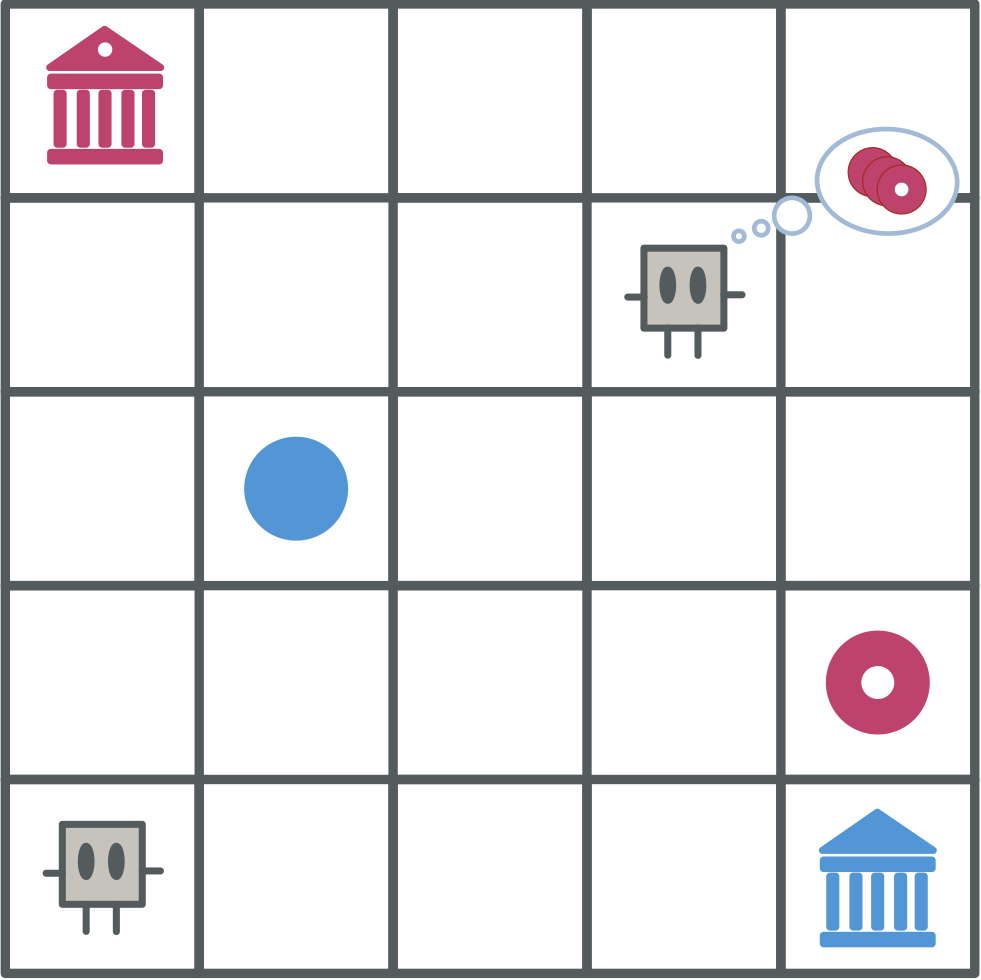}
        \caption{Environment (5x5)}
        \label{fig:treasure:game}
    \end{subfigure}
    \begin{subfigure}[b]{.55\columnwidth}
            \centering
            \includegraphics[width=\columnwidth]{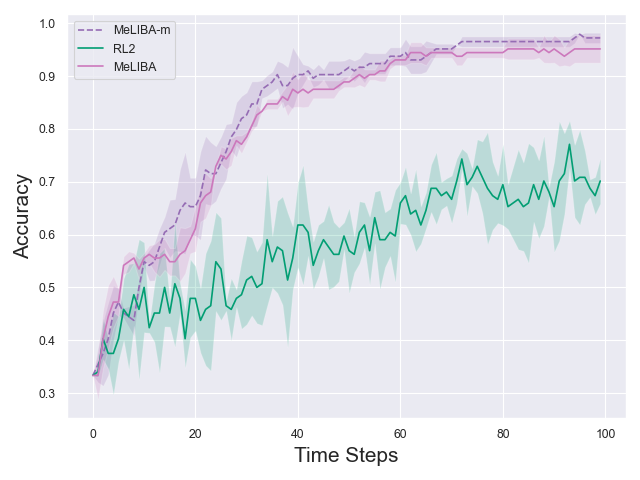}
            \caption{Agent Type Prediction Accuracy}
            \label{fig:treasure:accuracies}
    \end{subfigure}
\caption{Treasure Hunt Game.}
\label{fig:treasure_hunt}
\end{minipage}
\end{figure}

To this end, we use a gridworld version of the Treasure Hunt by \cite{iqbal2018actor}, which we designed to be able to easily control the other agents' strategy.
It is a collaborative game with two agents who have to collect coloured coins and bring them to banks (see Fig \ref{fig:treasure:game}).
Agents get a small bonus for collecting coins ($0.1$), a large bonus for dropping them at the correctly coloured bank ($1$), and a penalty for dropping them at the wrong bank ($-1$).
Coins re-spawn at random locations after being dropped at a bank.
We hard-code $3$ other agents: two that only collect coins of one colour (unless the agent accidentally picks up a different-coloured coin, in which case it brings the coin to the correct bank), and an agent which alternates between colours.
We use a $10\times10$ grid with horizon $100$.

If no coin of the preferred colour is available, it does nothing and waits.
To maximise return it is therefore beneficial to identify which colour the other agent prefers and focus on the other coins.
I.e., the agent has to learn to infer the other agent's type, learn how this translates to what the other agent will do, and use this information to adapt accordingly and focus on the coins of a different colour.

The results are shown in Table \ref{fig:treasure:accuracies} (averages across $3$ seeds and $95\%$ confidence intervals in brackets; learning curves are in the appendix, Figure \ref{fig:appendix:treasure_learning_curve}).

\textbf{Comparison to models with permanent latent variables.}
We first consider the comparison between \acro~and the two models \acro-m and LIOM, which use a permanent latent variable only. 
The results show that there is no significant difference between these methods.
At first glance this seems surprising given that one of the other agents switches between coin colours. 
However in this case, the VAE loss is lowest if the latent variable represents the agent type and the last coin collected, in which case the reconstruction term can be correct until the next coin is collected (and wrong only for terms further into the future).
So despite the model mismatch, the information that is most relevant for the policy can still be captured in the latent variable. 
This suggests that in practice, if we know upfront that the other agent's actions do not depend on the history, we can use the simpler model \acro-m. 
If this is not known, we do not lose out on performance if we use the full \acro~model.
LIOM performs as well as \acro-m in this setting, which indicates that sampling does not have any significant effect on performance. 

\textbf{Comparison to model-free methods.}
We first consider the comparison to the model-free method RL$^2$.
RL$^2$ learns to adapt to some extent and performs better than the average policy; however, it is significantly outperformed by \acro.
This failure of RL$^2$ to learn the task compared to \acro~requires us to take a closer look at what both models do, in order to understand where this difference come from. 
To this end, we train a logistic regression classifier (see Appendix \ref{appendix:treasure_hunt}) to predict the other agent's type from the latent states of the models: for \acro~that is $m$ and $m_t$, and for RL$^2$ that is the hidden state of the RNN.
The results are shown in Figure \ref{fig:treasure:accuracies}.
For RL$^2$, we see that the other agent's strategy is represented to some extent in the hidden state of the recurrent model, and that the accuracy goes up as the policy gathers more data.
Given that RL$^2$ fails to outperform the feed-forward policy however, it might not have learned to use the information about the other agent's strategy to adapt.
In contrast, we can accurately predict the other agent's type from the latent variables in \acro, confirming that predicting future actions of the other agent is a useful auxiliary task for learning to infer agent type in this environment.

\section{Discussion \& Conclusions} \label{sec:discussion}

We introduced \acro, a general method for meta-learning approximately Bayes-optimal policies for a given distribution of other agents that can scale to complex multi-agent environments.
We showed that \acro's model of other agents learns a hierarchical latent representation of other agent types, separating the permanent and temporal internal states. 
On several environments we demonstrated that learning this latent representation by predicting future actions of other agents, and conditioning the policy on it, allows an agent to adapt to others. 
We conclude that maintaining explicit beliefs over other agents helps compared to model-free approaches, especially when there is not a tight coupling between other agent type and rewards.

In order to learn more informative priors, the prior distribution could be learned from (human) expert data using imitation learning techniques \citep{song2018multi}, 
or by using existing multi-agent algorithms to pre-train the other agents \citep{canaan2020generating}. 
Interesting settings include humans playing video games together, where we might be interested in learning agents that can cooperate with humans in an ad-hoc setting;
or cars driving on the highway, to later train self-driving cars in simulation to navigate in traffic before deploying them in the real world.  

If such data is not available, another interesting direction for future work is to train all agents simultaneously, with the goal of generalising well to new agents that were not seen during training.
This is a difficult problem, since sufficient diversity has to be maintained and agents can overfit to the agents they have been trained with \citep{carroll2019utility, canaan2020evaluating}.
In addition, this introduces a different type of non-stationarity in the other agents which must be addressed, since all agents now also evolve in-between episodes.
In our current setting we assume the other agents can adapt \emph{within} a single episode, but are reset at the beginning of each episode (see Section \ref{sec:background:prior}).

Even with this assumption, computing a Bayes-optimal policy is challenging. 
If we additionally want to model other agents that learn over time (not just within a single episode), one way forward is to either assume their learning algorithm is known and use the interactions with them to do inference on how their character changes over time, or add a third level to our hierarchy (on top of the agents' permanent type $m$ and its temporal type $m_t$ that allows it to adapt within an episode), e.g., $m_l$, that models their latent \emph{learning} algorithm. 
Our agent would then have to reason about how its own actions and learning behaviour influences the other agents' learning behaviour over time. 
This is an incredibly challenging problem that we leave to future work. 

Other interesting future directions are to, e.g., not require access of the other agent's observations at test time \citep{papoudakis2020variational}, or include reward or state uncertainty \citep{gmytrasiewicz2005framework, ng2012bayes, oliehoek2014best}.


\begin{acks}
We thank Cheng Zhang, Yingzhen Li, Mikhail Jacob, Robert Loftin, Tabish Rashid, and everyone at Project Paidia (Microsoft Research) for useful discussions and feedback.
Luisa Zintgraf is also supported by the 2017 Microsoft Research PhD Scholarship Program, and the 2020 Microsoft Research EMEA PhD Award.
This project has received funding from the European Research Council under the European Union’s Horizon 2020 research and innovation programme (grant agreement number 637713).
\end{acks}


\bibliographystyle{plainnat}
\bibliography{references}


\newpage
\appendix
{\center \textbf{Appendix}}

\section{ELBO Derivation} \label{appendix:elbo_derivation}

We derive the ELBO for modelling the future actions of agent $i$ as follows.

\begin{align}
    & \log ~ p(u^{(i)}_{t:H}|s_t) \\
    & = \log \int p(u^{(i)}_{t:H}, m, m_t| s_t) \frac{q(m, m_t|\tau_{:t})}{q(m, m_t|\tau_{:t})} ~ d m ~ d m_t  \\
    & = \log \mathbb{E}_{q(m, m_t|\tau_{:t})} \bigg[ \frac{p(u^{(i)}_{t:H}, m, m_t|s_t)}{q(m, m_t|\tau_{:t})} \bigg] \\
    & \ge \mathbb{E}_{q(m, m_t|\tau_{:t})} \bigg[ \log \frac{p(u^{(i)}_{:H}, m, m_t|s_t)}{q(m, m_t|\tau_{:t})} \bigg]\\
    & = \mathbb{E}_{q(m, m_t|\tau_{:t})} \bigg[ \log p(u^{(i)}_{t:H}, m, m_t |s_t) -  \log q(m, m_t|\tau_{:t}) \bigg] \\
    & = \mathbb{E}_{q(m, m_t|\tau_{:t})} \bigg[ \log p(u^{(i)}_{t:H}| s_t, m, m_t ) + \log p(m, m_t|s_t) \nonumber \\
    & \hspace{2cm} -  \log q(m, m_t|\tau_{:t}) \bigg] \\
    & = \mathbb{E}_{q(m, m_t|\tau_{:t})} \bigg[ \log p(u^{(i)}_{t:H} | s_t, m, m_t) \bigg] \nonumber \\
    & \hspace{0.5cm} - KL(q(m, m_t|\tau_{:t})||p(m, m_t|s_t))
    .
\end{align}

\newpage 
We expand the left-hand side:
\begin{align}
    \mathbb{E}&_{q(m, m_t|\tau_{:t})} \bigg[ \log p(u^{(i)}_{t:H} | s_t, m, m_t) \bigg] \nonumber \\ 
    & = \mathbb{E}_{q(m, m_t|\tau_{:t})} \bigg[ \log \iint p(u_t^{(i)}, \mathbf{u}^{(-i)}_t, s_{t+1}, u_{t+1:H} | s_t, m, m_t) \nonumber \\
    & \hspace{3cm} d \mathbf{u}^{(-i)}_t d s_{t+1} \bigg] \label{eq:line:time_expansion_1} 
    \\ 
    & = \mathbb{E}_{q(m, m_t|\tau_{:t})} \bigg[ \log \iint 
    p( u^{(i)}_{t+1:H} |  u^{(i)}_t, \mathbf{u}^{(-i)}_t, s_{t+1}, s_t,  m, m_t). 
    \nonumber \\ 
    & 
    \hspace{3cm} p(u^{(i)}_t, \mathbf{u}^{(-i)}_t , s_{t+1}| s_t, m, m_t ) 
    \nonumber \\
    & \hspace{3cm} d \mathbf{u}^{(-i)}_t d s_{t+1} \bigg] \\
    & = \mathbb{E}_{q(m, m_t|\tau_{:t})} \bigg[ \log \iint 
    p( u^{(-i)}_{t+1:H} | u^{(i)}_t, \mathbf{u}^{(-i)}_t, s_{t+1}, s_t, m, m_t)  
    \nonumber \\ 
    &  \hspace{3cm} p(\mathbf{u}^{(-i)}_t, s_{t+1} | u^{(i)}_t, s_t, m, m_t )  
    \nonumber \\
    &  \hspace{3cm} p(u^{(i)}_t | s_t, m, m_t )  
    \nonumber \\
    & \hspace{3cm}
    d \mathbf{u}^{(-i)}_t  d s_{t+1} \bigg] 
    \\
    & = \mathbb{E}_{q(m, m_t|\tau_{:t})} \bigg[ \log \iint 
    p( u^{(-i)}_{t+1:H} | s_{t+1}, m, m_{t+1})  
    \nonumber \\
    &  \hspace{3cm} p(\mathbf{u}^{(-i)}_t, s_{t+1} | u^{(i)}_t, s_t, m, m_t )  
    \nonumber \\
    &  \hspace{3cm} p(u^{(i)}_t | s_t, m, m_t )  
    \nonumber \\
    & \hspace{3cm}
    d \mathbf{u}^{(-i)}_t  d s_{t+1} \bigg] 
    \\
    & = \mathbb{E}_{q(m, m_t|\tau_{:t})} \bigg[ \log 
    \mathbb{E}_{p(\mathbf{u}^{(-i)}_t, s_{t+1} | u^{(i)}_t, s_t, m, m_t )} 
    \bigg[ \nonumber \\ 
    & \hspace{1.5cm}
    p(u^{(-i)}_{t+1:H} | s_{t+1}, m, m_{t+1}) p(u^{(i)}_t | s_t, m, m_t )  
    \bigg] \bigg] \label{eq:line:timestep_iter_1} \\
    & \vdots \nonumber \\
    & = \mathbb{E}_{q(m, m_t|\tau_{:t})} \bigg[ \log 
    \mathbb{E}_{p(\mathbf{u}^{(-i)}_{t:H}, s_{t+1:H+1} | u^{(i)}_t, s_t, m, m_t )} 
    \bigg[ \nonumber \\ 
    & \hspace{3.3cm} \prod_{i=t+1}^H p(u^{(i)}_i | s_i, m, m_i )  
    \bigg] \bigg] 
    \label{eq:line:timestep_iter_2}
\end{align}

Going from \eqref{eq:line:timestep_iter_1} to \eqref{eq:line:timestep_iter_2}, we iterate over timesteps, repeating steps \eqref{eq:line:time_expansion_1}-\eqref{eq:line:timestep_iter_1} for $t+1,\dotsc,H$.

For right-hand side, following \cite{zintgraf2020varibad} we use the previous posterior as the new prior, $q(m, m_t|\tau_{:t-1})$, with univariate Gaussian priors $q(m, m_t|s_0)=\mathcal{N}(0,1)$ at the first timestep.

\newpage


\section{Experiment Details \& \\ Additional Results} \label{appendix:experiments}

\subsection{Game of Chicken} \label{appendix:results:chicken}

\begin{figure}
\begin{subfigure}{\columnwidth}
    \centering
    \includegraphics[width=\columnwidth]{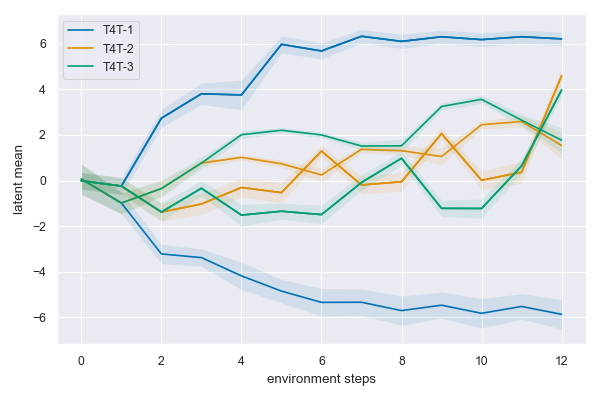}
    \caption{\acro-$m$}
    \label{fig:appendix:chicken_rollouts:meliba_m}
\end{subfigure}
\begin{subfigure}{\columnwidth}
    \centering
    \includegraphics[width=\columnwidth]{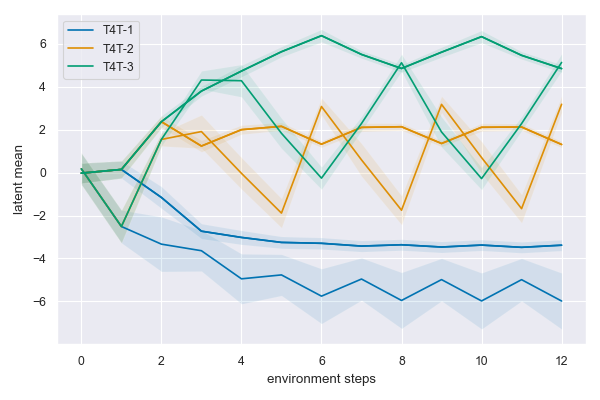}
    \caption{\acro-$m_t$}
    \label{fig:appendix:chicken_rollouts:meliba_mt}
\end{subfigure}
\begin{subfigure}{\columnwidth}
    \centering
    \includegraphics[width=\columnwidth]{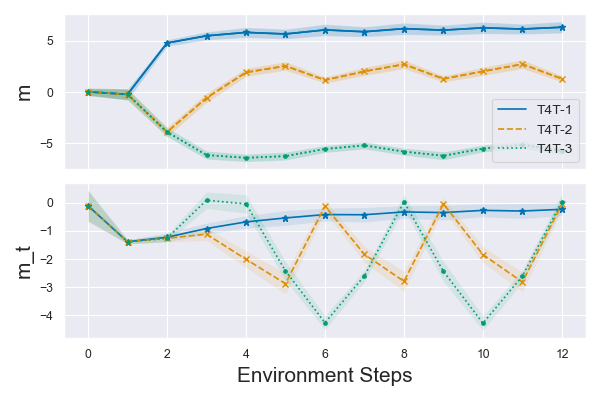}
    \caption{\acro-flat}
    \label{fig:appendix:chicken_rollouts:meliba_flat}
\end{subfigure}
\caption{Latent variables during test rollout, for different \acro~architecture ablations.}
\label{fig:appendix:chicken_rollouts}
\end{figure}

Figure \ref{fig:appendix:chicken_rollouts} shows the latent variables for different ablations of the \acro~architecture.
Figures \ref{fig:appendix:chicken_rollouts:meliba_m} and \ref{fig:appendix:chicken_rollouts:meliba_mt} show the latent variable per agent type, when using either only the permanent latent variable $m$ (together with a feed-forward decoder) or only the temporal latent variable $m_t$ (together with a recurrent decoder).
We first observe that both seem to ``count'' the number of interactions to some extend, although the permanent latent $m$ does so less pronounced and consistent. 
Figure \ref{fig:appendix:chicken_rollouts:meliba_flat} shows the latent variables at test time, when using a non-hierarchical version of \acro. We can see that some of the temporal structure is also captured in $m$ here. 
This is undesirable, because we want the model to clearly separate temporal and permanent structure (like \acro~does, see Figure \ref{fig:chicken_results:latent_visualisation}).

\begin{figure}
    \centering
    \includegraphics[width=\columnwidth]{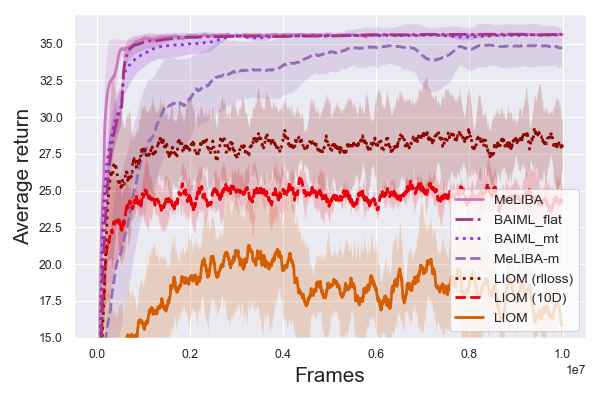}
    \caption{Additional Learning Curves for the Chicken Game.}
    \label{fig:appendix:chicken_learning_curve}
\end{figure}

Figure \ref{fig:appendix:chicken_learning_curve} shows additional learning curves for variations of LIOM and ablations of \acro.
We see that all \acro~architectures learn to solve the task, except when we use only a permanent latent variable $m$.
The reason is a mismatch in model type: the other agent conditions its actions on the interaction history, however, the action decoder is not able to model this since it is a feed-forward network conditioned only on the last state and actions.
We also see that LIOM's \citep{papoudakis2020variational} performance increases when either backpropagating the RL loss through the encoder (LIOM (rlloss)), or when using a larger latent dimension (LIOM (10D)). 
Still, the performance is inferior in both these cases. 
Since the only difference between \acro~and LIOM is that LIOM passes a sample to the policy instead of the mean and variance, we believe that the noisy inputs this generates for the policy lead to an instability in training.

\subsection{Treasure Hunt} \label{appendix:treasure_hunt}

We use a $10\times10$ grid and the agents can choose from five discrete actions \emph{no-op, left, right, up, down}.
The rewards are $-0.1$ if the agents collide, $+0.1$ for picking up a coin, $+1$ for dropping off a coin at the correct bank, and $-1$ for dropping off a coin at the wrong bank. If the agents attempt to walk into a wall or another agent, their action does not get executed and they stay on the current grid. State transitions are deterministic.

The hard-coded agents go directly to the coin colour that they prefer, without avoiding other-coloured coins or the other agent.
If they pick up a coin of the wrong colour, they drop it off at the correct bank and then proceed with their favourite coin colour.

\begin{figure}
    \centering
    \includegraphics[width=\columnwidth]{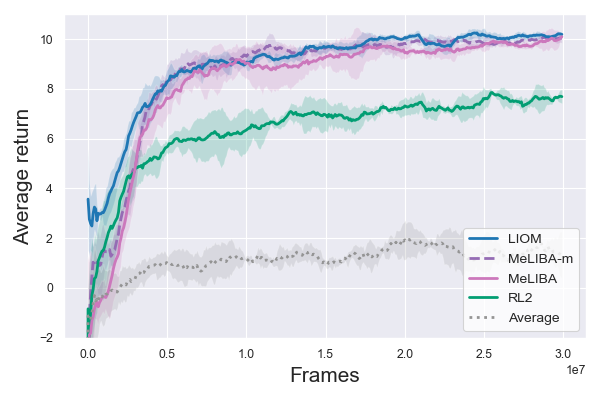}
    \caption{Treasure Hunt Learning Curves.}
    \label{fig:appendix:treasure_learning_curve}
\end{figure}

\textbf{Learning Curves.}
Figure \ref{fig:appendix:treasure_learning_curve} shows the learning curves for the Treasure Hunt game, for $3$ seeds per method. 
All learning curves show a $95\%$ confidence intervals in the shaded areas, across the different seeds.

\textbf{Classification.} 
To classify agent type from latent variables, we fit a Logistic Regression Classifier using the scikit-learn Python package. Per trained model, We use 80 test rollouts for fitting the classifier, and show the test accuracy on $20$ rollouts in Figure \ref{fig:treasure:accuracies}. The shaded areas are one standard deviation across $3$ trained models (different seeds).

\subsection{Runtimes}

We trained our models using Nvidia Tesla K80 GPUs.
Table \ref{table:runtimes} shows the runtimes for the different methods.
We typically ran between 1 and 4 experiments per GPU in parallel, leading to some variation in the runtime. 
Hence these numbers should be seen as a rough guide and are useful to compare relative times between methods.

\begin{table}[]
\centering
\begin{tabular}{|l|c|c|l}
\cline{1-3}
\multicolumn{1}{|c|}{Method} 
& \begin{tabular}[c]{@{}c@{}}Chicken Game\\ (10M Frames)\end{tabular} 
& \begin{tabular}[c]{@{}c@{}}Treasure Hunt\\ (25M Frames)\end{tabular} 
&  
\\ \cline{1-3}
MeLIBA                        
& 2-3                                                                 
&5                                                                       
&  
\\ \cline{1-3}
MeLIBA (m)                    
& 1-2                                                                 
& 5                                                                     
&  
\\ \cline{1-3}
RL$^2$                       
& 4                                                                   
& 12                                                                     
&  
\\ \cline{1-3}
Feedforward                  
& 1                                                                   
& 3                                                                       
&  
\\ 
\cline{1-3}
\end{tabular}
\caption{Runtimes in Hours}
\label{table:runtimes}
\end{table}

\subsection{Hyperparameters} \label{appendix:hyperparameters}

Hyperparameters for \acro~are shown in Table \ref{table:hyperparameters}. 

Additional Notes:
\begin{itemize}
    \item $^*$: In the first layer of the encoder, we separately encode the state/action/rew with a FC layer with output $32/16/16$ for the Chicken Game, and $64/16/16$ for the Treasure Hunt.
    \item $^R$: Denotes the (hidden size of the) recurrent layer.
    \item To Train $RL^2$, we used a recurrent network with hidden size $128$ and three layers.
    \item We use GRUs as our aggregator in the recurrent parts of the networks.
    \item We selected the hyperparameters doing a simple linesearch (see Table for the values we considered).
\end{itemize}

\begin{table*}[]
\begin{tabular}{llccc}
%
\hline
\multicolumn{2}{|c|}{\multirow{2}{*}{\textbf{Parameter}}}                                    
& \multicolumn{2}{c|}{\textbf{Values Chosen}}                                                                        
& \multicolumn{1}{c|}{\multirow{2}{*}{\textbf{\begin{tabular}[c]{@{}c@{}}Values \\ Tested\end{tabular}}}} 
\\ \cline{3-4}
\multicolumn{2}{|c|}{}                                                                       
& \multicolumn{1}{c|}{Chicken Game}                                                                 
& \multicolumn{1}{c|}{Treasure Hunt}                                                                 
& \multicolumn{1}{c|}{}                                                                                   
\\ \hline \hline
%
\multicolumn{2}{|l|}{Num Training Frames}                                                    
& \multicolumn{1}{c|}{1e7}                                                                          
& \multicolumn{1}{c|}{2.5e7}                                                                         
& \multicolumn{1}{c|}{}                                                                                   
\\ \hline
\multicolumn{1}{|l|}{\multirow{14}{*}{Policy}} 
& \multicolumn{1}{l|}{Algorithm}              
& \multicolumn{1}{c|}{PPO}                                                                          
& \multicolumn{1}{c|}{PPO}                                                                           
& \multicolumn{1}{c|}{}                                                                                   
\\ \cline{2-5} 
\multicolumn{1}{|l|}{}                         
& \multicolumn{1}{l|}{Layers}                 
& \multicolumn{1}{c|}{[128, 128]}                                                                   
& \multicolumn{1}{c|}{[128, 128]}                                                                    
& \multicolumn{1}{c|}{}                                                                                   
\\ \cline{2-5}
\multicolumn{1}{|l|}{}                         & \multicolumn{1}{l|}{Non-Linearity}          & \multicolumn{1}{c|}{tanh}                                                                         & \multicolumn{1}{c|}{tanh}                                
& \multicolumn{1}{c|}{}                                                                                   \\ \cline{2-5} 
\multicolumn{1}{|l|}{}                         & \multicolumn{1}{l|}{Optimiser}              & \multicolumn{1}{c|}{RMSProp}                                                                      & \multicolumn{1}{c|}{RMSProp}        
& \multicolumn{1}{c|}{Adam}                                                                               \\ \cline{2-5} 
\multicolumn{1}{|l|}{}                         & \multicolumn{1}{l|}{Learning Rate}          & \multicolumn{1}{c|}{7e-4}                                                                         & \multicolumn{1}{c|}{7e-4}       
& \multicolumn{1}{c|}{1e-4, 1e-3}                                                                         \\ \cline{2-5} 
\multicolumn{1}{|l|}{}                         & \multicolumn{1}{l|}{Num Epochs}             & \multicolumn{1}{c|}{2}                                                                            & \multicolumn{1}{c|}{2}         
& \multicolumn{1}{c|}{1, 2, 4}                                                                            \\ \cline{2-5} 
\multicolumn{1}{|l|}{}                         & \multicolumn{1}{l|}{Num Mini-Batches}       & \multicolumn{1}{c|}{8}                                                                            & \multicolumn{1}{c|}{4}          
& \multicolumn{1}{c|}{2, 4, 8}                                                                            \\ \cline{2-5} 
\multicolumn{1}{|l|}{}                         & \multicolumn{1}{l|}{Value Clip Param}       & \multicolumn{1}{c|}{0.1}                                                                          & \multicolumn{1}{c|}{0.1}      
& \multicolumn{1}{c|}{}                                                                                   \\ \cline{2-5} 
\multicolumn{1}{|l|}{}                         & \multicolumn{1}{l|}{Batchsize}              & \multicolumn{1}{c|}{2,080}                                                                        & \multicolumn{1}{c|}{1,600}    
& \multicolumn{1}{c|}{}                                                                                   \\ \cline{2-5} 
\multicolumn{1}{|l|}{}                         & \multicolumn{1}{l|}{Value Loss Coefficient} & \multicolumn{1}{c|}{0.5}                                                                          & \multicolumn{1}{c|}{0.5}      
& \multicolumn{1}{c|}{}                                                                                   \\ \cline{2-5} 
\multicolumn{1}{|l|}{}                         & \multicolumn{1}{l|}{Entropy Coefficient}    & \multicolumn{1}{c|}{0.2}                                                                          & \multicolumn{1}{c|}{0.2}      
& \multicolumn{1}{c|}{0.1, 0.5}                                                                           \\ \cline{2-5} 
\multicolumn{1}{|l|}{}                         & \multicolumn{1}{l|}{Discount Factor}        & \multicolumn{1}{c|}{1}                                                                            & \multicolumn{1}{c|}{1}        
& \multicolumn{1}{c|}{}                                                                                   \\ \cline{2-5} 
\multicolumn{1}{|l|}{}                         & \multicolumn{1}{l|}{Tau (for GAE)}          & \multicolumn{1}{c|}{0.9}                                                                          & \multicolumn{1}{c|}{0.9}      
& \multicolumn{1}{c|}{0.95}                                                                               \\ \cline{2-5} 
\multicolumn{1}{|l|}{}                         & \multicolumn{1}{l|}{Max Grad Norm}          & \multicolumn{1}{c|}{0.5}                                                                          & \multicolumn{1}{c|}{0.5}      
& \multicolumn{1}{c|}{}                                                                                   \\ \hline
\multicolumn{1}{|l|}{\multirow{11}{*}{VAE}}    & \multicolumn{1}{l|}{Learning Rate}          & \multicolumn{1}{c|}{0.001}                                                                        & \multicolumn{1}{c|}{0.001}    
& \multicolumn{1}{c|}{}                                                                                   \\ \cline{2-5} 
\multicolumn{1}{|l|}{}                         & \multicolumn{1}{l|}{Optimiser}              & \multicolumn{1}{c|}{Adam}                                                                         & \multicolumn{1}{c|}{Adam}       
& \multicolumn{1}{c|}{}                                                                                   \\ \cline{2-5} 
\multicolumn{1}{|l|}{}                         & \multicolumn{1}{l|}{Data Buffer}            & \multicolumn{1}{c|}{2500 (unique traj)}                                                           & \multicolumn{1}{c|}{2500}    
& \multicolumn{1}{c|}{10000}                                                                              \\ \cline{2-5} 
\multicolumn{1}{|l|}{}                         & \multicolumn{1}{l|}{Batchsize (Num Traj)}   & \multicolumn{1}{c|}{50}                                                                           & \multicolumn{1}{c|}{15}      
& \multicolumn{1}{c|}{}                                                                                   \\ \cline{2-5} 
\multicolumn{1}{|l|}{}                         & \multicolumn{1}{l|}{KL weight}              & \multicolumn{1}{c|}{0.05}                                                                         & \multicolumn{1}{c|}{0.05}    
& \multicolumn{1}{c|}{0, 0.01, 0.1}                                                                       \\ \cline{2-5} 
\multicolumn{1}{|l|}{}                         & \multicolumn{1}{l|}{Pretrain: Num Frames}   & \multicolumn{1}{c|}{208,000}                                                                      & \multicolumn{1}{c|}{32,000}  
& \multicolumn{1}{c|}{}                                                                                   \\ \cline{2-5} 
\multicolumn{1}{|l|}{}                         & \multicolumn{1}{l|}{Pretrain: Num Updates}  & \multicolumn{1}{c|}{5000}                                                                         & \multicolumn{1}{c|}{1000}    
& \multicolumn{1}{c|}{}                                                                                   \\ \cline{2-5} 
\multicolumn{1}{|l|}{}                         & \multicolumn{1}{l|}{Latent dim, m}          & \multicolumn{1}{c|}{2}                                                                            & \multicolumn{1}{c|}{5}       
& \multicolumn{1}{c|}{2, 5, 10}                                                                           \\ \cline{2-5} 
\multicolumn{1}{|l|}{}                         & \multicolumn{1}{l|}{Latent dim, $m_t$}      & \multicolumn{1}{c|}{2}                                                                            & \multicolumn{1}{c|}{5}       
& \multicolumn{1}{c|}{2, 5, 10}                                                                           \\ \cline{2-5} 
\multicolumn{1}{|l|}{}                         & \multicolumn{1}{l|}{Encoder Layers}         & \multicolumn{1}{c|}{\begin{tabular}[c]{@{}c@{}}$[64^*, 64^R,$ \\ $64, 64^{m_t}]$\end{tabular}}    & \multicolumn{1}{c|}{\begin{tabular}[c]{@{}c@{}}$[96^*, 128^R, $\\ $64, 64^{m_t}]$\end{tabular}}   
& \multicolumn{1}{c|}{}                                                                                   \\ \cline{2-5} 
\multicolumn{1}{|l|}{}                         & \multicolumn{1}{l|}{Decoder Layers}         & \multicolumn{1}{c|}{\begin{tabular}[c]{@{}c@{}}$[32^m, 64^{m_t},$\\ $64, 64^R, 32]$\end{tabular}} & \multicolumn{1}{c|}{\begin{tabular}[c]{@{}c@{}}$[32^m, 64^{m_t},$\\ $64, 64^R, 32]$\end{tabular}} 
& \multicolumn{1}{c|}{}                                                                                   \\ \hline
                                               &                                             & \multicolumn{1}{l}{}                                                                              & \multicolumn{1}{l}{}                                                                                   
                                               & \multicolumn{1}{l}{}                                                                                   
\end{tabular}
\caption{\acro~Hyperparameters}
\label{table:hyperparameters}
\end{table*}


\end{document}